\documentclass[sigconf]{acmart}
\usepackage{booktabs}    
\usepackage{pdfpages}    
\usepackage{amsmath, bm} 
\usepackage{textcomp}
\usepackage{multirow}
\usepackage{multicol}
\usepackage{makecell}
\usepackage{relsize}
\usepackage{siunitx}
\usepackage{numprint}
\usepackage[utf8]{inputenc}

\acmSubmissionID{112}

\newlength{\Oldarrayrulewidth}

\copyrightyear{2021}
\acmYear{2021}
\setcopyright{rightsretained}
\acmConference[SA '21 Technical Communications]{SIGGRAPH Asia 2021 Technical Communications}{December 14--17, 2021}{Tokyo, Japan}
\acmBooktitle{SIGGRAPH Asia 2021 Technical Communications (SA '21 Technical Communications), December 14--17, 2021, Tokyo, Japan}
\acmDOI{10.1145/3478512.3488599}
\acmISBN{978-1-4503-9073-6/21/12}

\setcitestyle{square}

\makeatletter
\let\@authorsaddresses\@empty
\makeatother

\begin{document}

\title[TMT: A Data-Driven Approach for Versatile and Controllable Agents in Simulated Environments]{Transition Motion Tensor: A Data-Driven Approach for Versatile and Controllable Agents in Physically Simulated Environments}

\author{Jonathan Hans Soeseno}
\email{soeseno.jonathan@inventec.com}
\authornote{Joint first authors.}
\affiliation{\institution{Inventec Corp.}
            \city{Taipei}
            \country{Taiwan}}

\author{Ying-Sheng Luo}
\email{luo.ying-sheng@inventec.com}
\authornotemark[1]
\affiliation{\institution{Inventec Corp.}
             \city{Taipei}
             \country{Taiwan}}

\author{Trista Pei-Chun Chen}
\email{chen.trista@inventec.com}
\affiliation{\institution{Inventec Corp.}
                \city{Taipei}
                \country{Taiwan}}
                
\author{Wei-Chao Chen}
\email{chen.wei-chao@inventec.com}
\affiliation{\institution{Inventec Corp.}
                \city{Taipei}
                \country{Taiwan}}

\begin{abstract}
This paper proposes the Transition Motion Tensor, a data-driven framework that creates novel and physically accurate transitions outside of the motion dataset. It enables simulated characters to adopt new motion skills efficiently and robustly without modifying existing ones. Given several physically simulated controllers specializing in different motions, the tensor serves as a temporal guideline to transition between them. Through querying the tensor for transitions that best fit user-defined preferences, we can create a unified controller capable of producing novel transitions and solving complex tasks that may require multiple motions to work coherently. We apply our framework on both quadrupeds and bipeds, perform quantitative and qualitative evaluations on transition quality, and demonstrate its capability of tackling complex motion planning problems while following user control directives.
\end{abstract}




\begin{teaserfigure}
    \centering
    \vspace{-0.25cm}
    \includegraphics[width=6.8in]{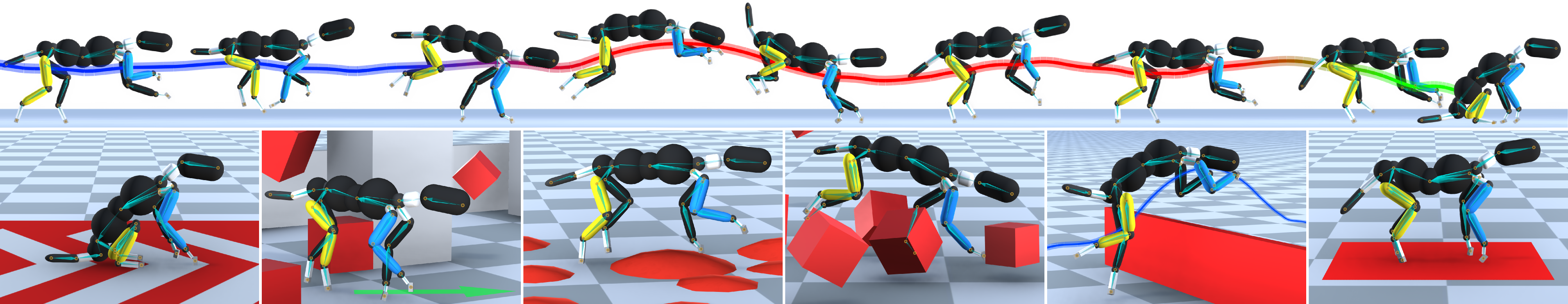}
    \caption{Our data-driven approach compiles individual controllers into a coherent unified controller.}
    \label{fig:teaser}
\end{teaserfigure}

\maketitle

\vspace{-0.2cm}
\section{Background}
The development of character controllers plays a crucial role in movies, games, and robotics. These applications have different requirements for the controller, such as high-fidelity movements that look natural and crisp for movies, responsiveness while maintaining some degree of realism for games, and prioritization of character's safety over motion quality for legged robots. However, controllers for these applications still share similar properties to enable a wide variety of character motions.

To achieve this, one may rely on capturing or authoring an extensive motion dataset to develop a character controller based on the recorded character joint's relations ~\cite{zhang2018mann, holden2017phase, starke2020localphase, Kovar2002motion}. While we can enrich the character's motion repertoire by collecting more data, these kinematic approaches require the motion dataset to capture sufficient environment variations to enable character-scene interaction, which could be infeasible to collect for complex interactions. 

Their counterpart, the physics-based approaches, can produce plausible scene interactions since they simulate character movements in a physics-enabled environment. Although the constraints from the environment can filter out physically infeasible movements, they also make the controller harder to train. For example, prior works from ~\citep{luo2020carl, peng2019mcp, bergamin2019drecon, liu2016guided, park2019learning, won2020scadiver} produce character controllers that are versatile and controllable but struggle to accommodate a more extensive motion vocabulary because of the requirement to re-train the respective modules when the vocabulary is expanded. Not only the re-training process takes an increasingly longer time with more motions, but it can also alter, or worse yet, break the existing motions.

To create a data-driven physics-based controller that scales with a large number of motions, we adopt the explicit controller assignment strategy to confine the motion complexity within individual controllers, called template controllers. We then observe that timing is crucial in performing transitions. Therefore, to reliably connect the controllers, we create an external module called Transition Motion Tensor (TMT) that applies to various characters (see Figure \ref{fig:teaser} and Figure \ref{fig:showcase_result}). We summarize our contribution as follows,

\begin{itemize}
    \item A data-driven approach to discover novel and physically accurate transitions outside the motion dataset,
    \item A scalable framework to link existing controllers of various architectures and training procedures into a coherent, unified controller without additional training, and
    \item A scheme to integrate user preferences, including the transition effort, duration, and control accuracy, into the controller's behavior.
\end{itemize}

\begin{figure}[t]
  \centering
  \includegraphics[width=.95\linewidth]{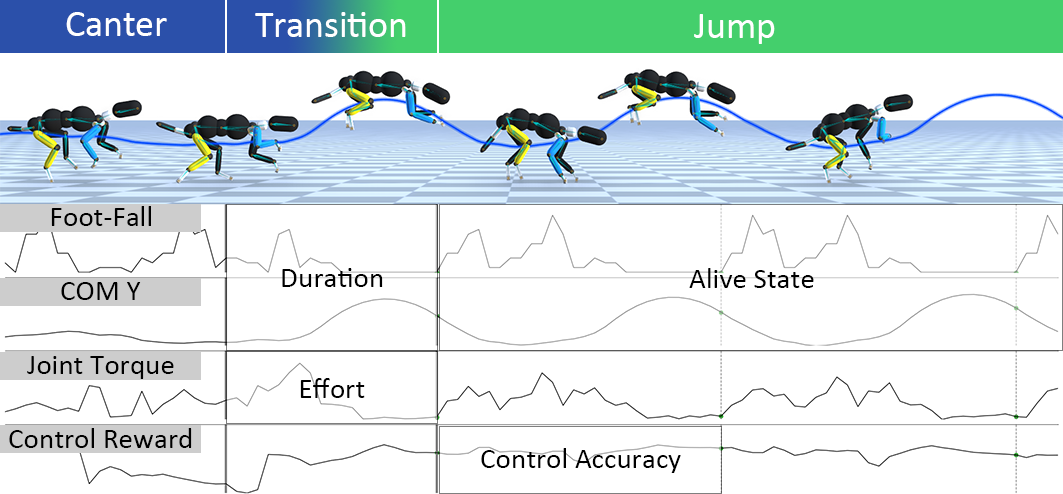}
  \caption{Transitioning from Canter to Jump, the destination controller stabilizes from the switching process, generating a transition of which we record the alive state ($\eta$), duration ($\Delta t$), effort ($e$), and control accuracy ($\alpha$) as the outcome.}
  \label{fig:transition_timeline}
    \vspace{-0.25cm}
\end{figure}
\section{Method}
Our goal is to enable a character to perform a wide array of motions in a physically simulated environment. To this end, we train controllable and robust template controllers, followed by unifying them using a data-driven tensor formulated based on user preferences.

\subsection{Template Controllers}
To achieve a versatile character,  we need to model the complex interaction between the desired motions. Instead of using a single controller to learn the entire motion vocabulary, which may require a substantial computation due to entangled motion complexity, we assign individual and physically simulated controllers to imitate specific motions, further referred to as the template controllers. Doing so allows us to confine each motion's complexity within each template controller, thereby making the training process more tractable and independent. We arrange each motion as a cycle, in which it starts and ends with a similar pose. Inspired by \citep{luo2020carl}, we train the template controllers using deep reinforcement learning to achieve life-like character movements, where the controller $\pi(\textbf{a}_t|\textbf{s}_t, \textbf{c}_t)$ yields the action $\textbf{a}_t$ at each timestep $t$ given the current state of the character $\textbf{s}_t$ and the control directives $\textbf{c}_t$.

The template controllers can perform the motions repeatedly in a cyclic manner, where similar character states re-appear every cycle. This condition indicates that the controllers are currently in a \emph{stable state}. However, since the character can interact with the environment, it may stumble due to external perturbations, causing it to break out from the cycle into an \emph{unstable state}. Therefore, to further ensure the robustness of the controllers, we introduce external perturbations and ask the controllers to stabilize the character back towards the cycle. The output is several controllable and robust template controllers $\Psi = \{\pi_1 ... \pi_6\}$, each specializing on a specific motion in the vocabulary $V=\{Trot, Canter, Jump, Pace, Stand, Sit\}$.

\begin{figure}[t]
  \centering
  \includegraphics[width=.95\linewidth]{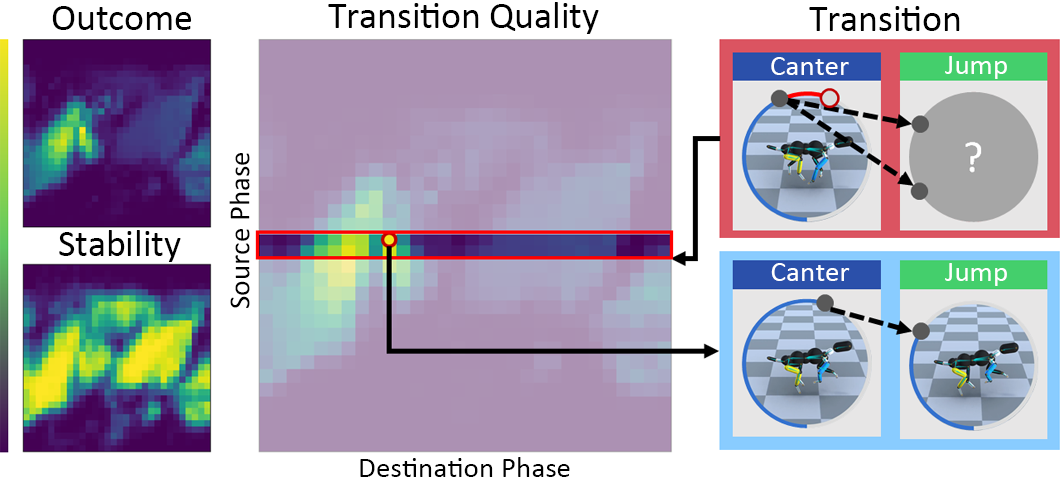}
  \vspace{-0.25cm}
  \caption{We express the quality of the transition by considering the transition outcome and stability. Querying for high-quality transitions, we can reliably transition between two controllers.}
  
  \label{fig:query_function}
  \vspace{-0.25cm}  
\end{figure}

\subsection{Transitioning between Controllers}
Each template controller only allows the character to perform a specific motion, limiting the character's ability to solve complex tasks that may require multiple motions to work coherently. For instance, to jump over a wide hole and quickly reach the destination, the character needs to run fast enough, followed by jumping and running towards the goal. However, knowing when to transition between running and jumping is not a trivial task since the character's state directly affects the transition outcome. 

Given that the character is in a particular state with the source controller, as we switch the controller from the source to the destination, the destination controller may have never seen this state. While it tries to recover from this unseen state (i.e., unstable), it consequently generates a new transition motion that neither exists in the source nor the destination controller. That is, the novel transitions are generated by switching between the pair of controllers at the appropriate timing. Without proper consideration of transition timing, we may introduce poses that are difficult for the destination controller to stabilize. For instance, the success of a transition between Canter and Jump relies heavily on the character's foot touching the ground. Therefore, transitioning from Jump to Canter when the character is mid-air may cause intricacies for the destination controller, leading to a longer time to stabilize, exerting too much effort, deviating from the control objectives, or even worse, causing the character to fall. 

To describe the likelihood of successful transitions between source and destination motions, we formulate a 4-D \emph{transition tensor} to record the outcomes of the transitions,

\begin{equation}
    \label{eq:transition_tensor}
    \nonumber
    \mathcal{T}_{m,\phi,n,\omega} = (\eta, \Delta t, e, \alpha),
\end{equation}
where $(\eta, \Delta t, e, \alpha)$ denotes the \emph{transition outcomes}. The indices of the tensor include the source $m \in V$ and the destination $n \in V$ motions, as well as the source $\phi \in [0, 1)$ and the destination phase $\omega \in [0, 1)$. Note that each component of the tensor is a 4-D vector $(\eta, \Delta t, e, \alpha)$ and should be dependent on $\textbf{w} = (m,\phi,n,\omega)$, e.g. $\eta \equiv \eta_\textbf{w}$, but we omit them for the sake of brevity. Thus, each element $\mathcal{T}_\textbf{w}$ represents the outcomes of transitions at $\textbf{w}$ where,
\begin{itemize}
    \item ($\eta$) denotes the \emph{alive state} after the transition, in which $\eta=1$ means successful transition and $\eta=0$ if the character falls,
    \item ($\Delta_t$) is the \emph{transition duration}, it starts with the switching process and ends after the destination controller stabilizes, 
    \item ($e$) expresses the \emph{effort} of a transition measured through summation of all joint's torque during the transition, and 
    \item ($\alpha$) denotes the \emph{control accuracy} measured by the sum of control rewards within the first cycle after the transition.
\end{itemize}

Figure \ref{fig:transition_timeline} provides a visual illustration of the transition outcomes.

\begin{figure}[t]
  \centering
  \includegraphics[width=.95\linewidth]{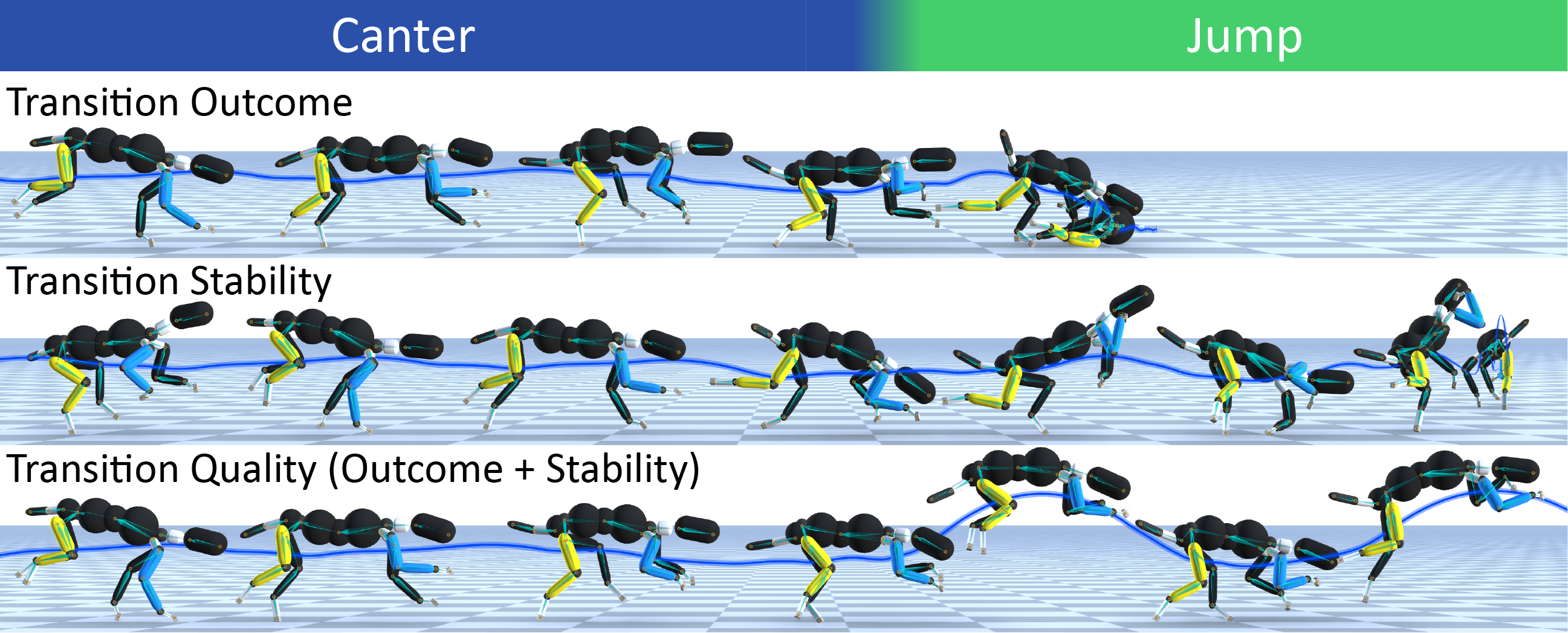}
  \caption{Excluding either the transition outcome or stability, the controller generates sub-optimal transitions.}
  \vspace{-0.3cm}
  \label{fig:ablation}
\end{figure}

\subsection{Unifying Controllers}
With the tensor $\mathcal{T}_{\textbf{w}}$ describing the likelihood of transitions between a set of template controllers, our goal is to unify all template controllers with transitions that are quick, obey control directives, require low effort, and most importantly, keep the character alive. Therefore, we consolidate the transition outcomes as follows,
\begin{equation}
    \nonumber
    \label{eq:transition_outcomes}
    \Gamma_\textbf{w} = \eta_\textbf{w} \times \frac{\alpha_\textbf{w}}{e_\textbf{w}} \times \exp(-\Delta t_\textbf{w}).
\end{equation} 

The tensor $\Gamma$ then denotes the consolidated outcome over the indices $\textbf{w}$. In addition to the outcomes, we also wish to ensure the transition stability where the controller can reliably generate good outcomes with high alive probability. Thus, we define a local neighborhood $\Gamma_\textbf{w}(\delta)$ which is a 2-D sub-tensor of $\Gamma$ near $\textbf{w}$, $\textbf{w} \in \{m,\phi\pm\delta,n,\omega\pm\delta\}$. Then, we can calculate the consistency of the transition outcome $\zeta_\textbf{w}(\delta)$ as the variance of all samples in $\Gamma_\textbf{w}(\delta)$.  Similarly, we compute the alive probability of a transition $\eta_\textbf{w}(\delta)$ as the proportion of samples within $\mathcal{T}_\textbf{w}(\delta)$ having $\eta = 1$. The final form of the transition stability then becomes,
\begin{equation}
    \nonumber
    \psi_\textbf{w}(\delta) = \eta_\textbf{w}(\delta) \times \exp(-\beta\zeta_\textbf{w}(\delta)), \beta = 0.015.
\end{equation}

Combining the transition stability and the outcomes, the quality of a transition at $\textbf{w}$ is,
\begin{equation}
    \label{eq:transition_quality}
    \nonumber
    \begin{aligned}
    \mathcal{Q}_\textbf{w} = \psi_\textbf{w}(\delta) \times \Gamma_\textbf{w}.
    \end{aligned} 
\end{equation}

Next, we look into the importance of each component of the transition quality. Figure \ref{fig:ablation} demonstrates the effect of removing either the transition outcome or stability. As expected, without considering the transition stability, the controllers generate transitions that cause the character to fall. On the other hand, ignoring the transition outcomes leads to awkward and inefficient transitions. Querying for the highest transition quality based on both components (see Figure \ref{fig:query_function}), we can then reliably perform transitions between two controllers. As a result, applying the same strategy for each pair-wise transition, we successfully unify all template controllers into a single versatile controller that can coherently perform all motions in the vocabulary. More visual illustration is available in the supplementary video.

\begin{figure}[t]
  \centering
  \includegraphics[width=.95\linewidth]{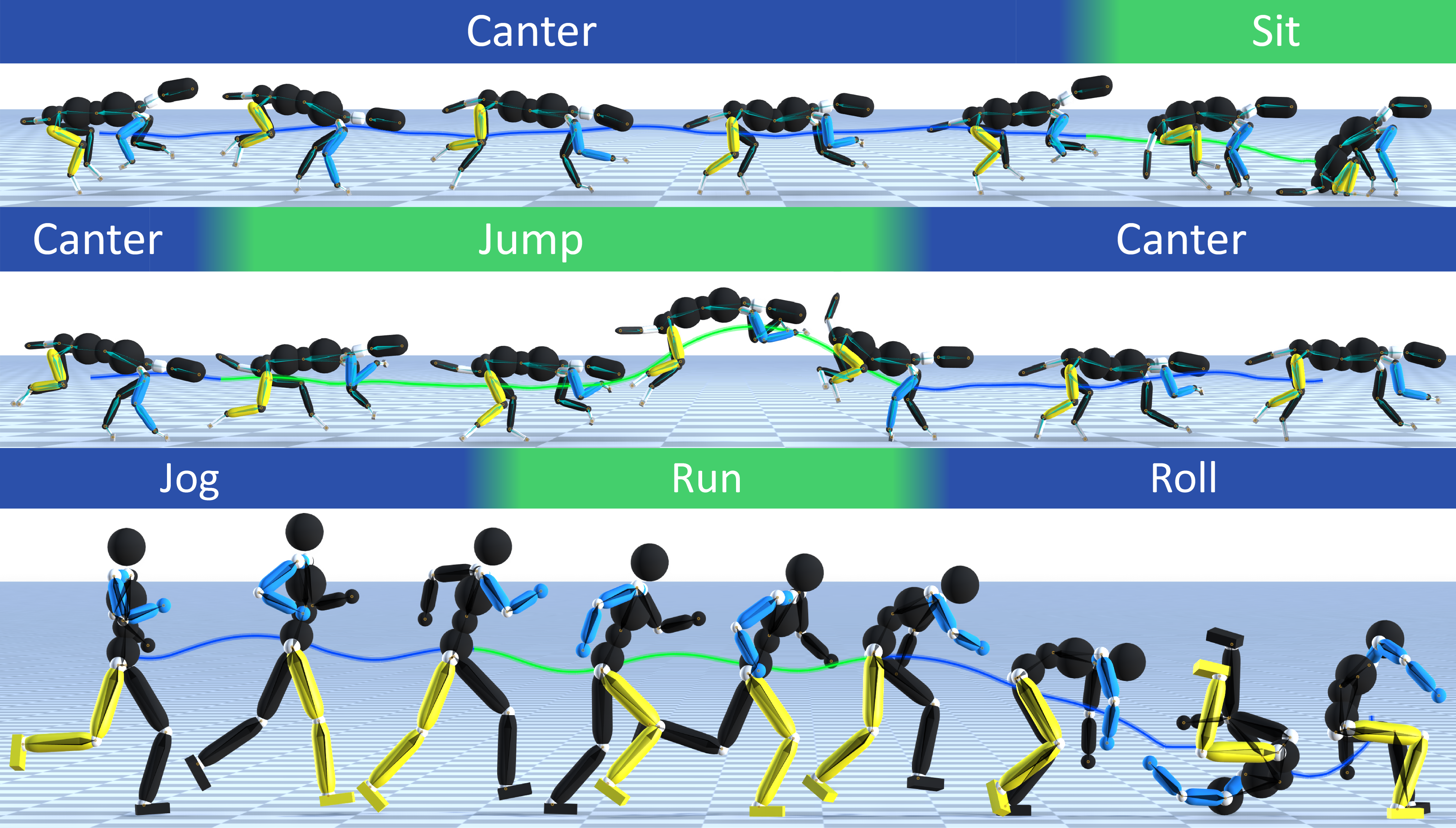}
  \caption{The transition tensor successfully identifies intricate transition strategies for quadruped and biped.}
  \label{fig:showcase_result}
  \vspace{-0.3cm}
\end{figure}

\section{Experiment}
This section evaluates the effectiveness of transition tensor in accommodating various motions in the vocabulary. 

\subsection{Comparison}
We compare our approach \textbf{TMT} to three other strategies. First, we consider performing \textbf{Motion Imitation}, where we ask a single template controller trained following \citep{luo2020carl} to imitate a reference motion clip containing all desired motions. Alternatively, we can train a \textbf{Gating Network} to additively compose the template controllers, as demonstrated by \citep{won2020scadiver}. Finally, inpsired by \citep{peng2018deepmimic}, we consider using each template controller's \textbf{Value Function} to regulate the transitions between them. We compare these strategies in a setting where we introduce an iteratively growing motion vocabulary using a single workstation equipped with 16 Cores of AMD EPYC 7702, with training speed averaging at 1.6e6 samples/h. Starting with Trot and Canter in the motion vocabulary, we expand it by adding a Jump motion. While the order of motion inclusion may affect the performance, the relative rank of the strategies remains similar. 

\begin{figure}[t]
  \centering
  \includegraphics[width=.95\linewidth]{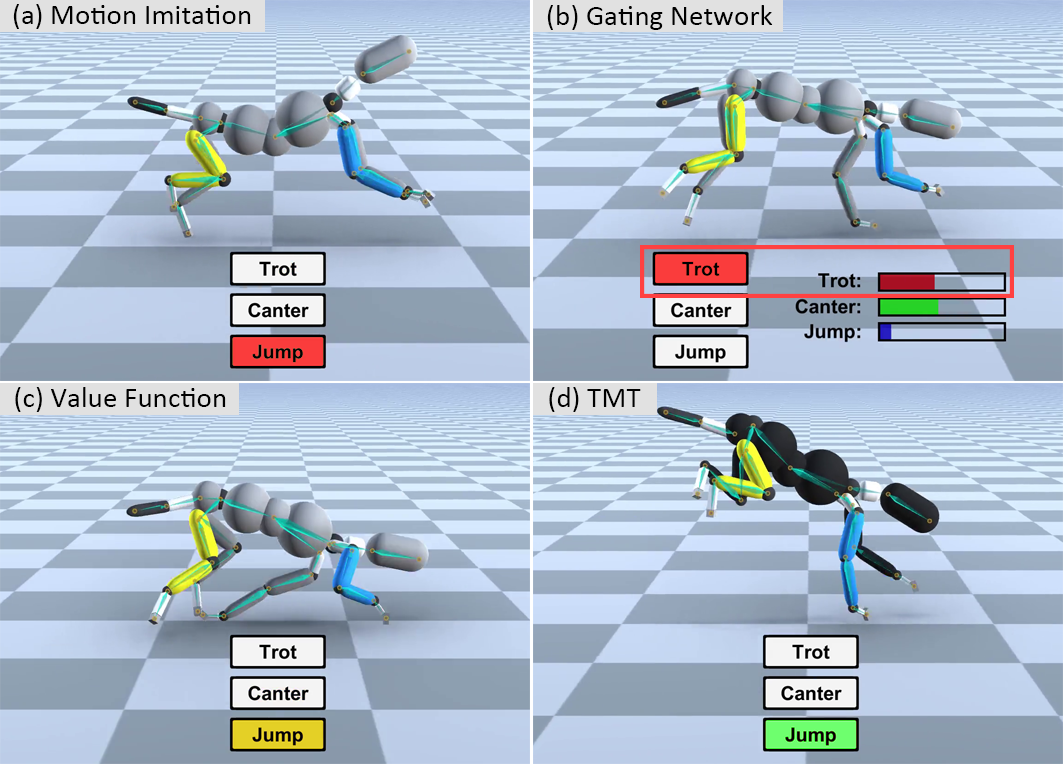}
  \caption{(a) Motion Imitation struggles to learn the Jump motion; (b) Gating Network strategy forgets the existing motion; (c) Value Function generates unstable transitions; and (d) TMT successfully adopts Jump and generates reliable transitions without altering Trot and Canter.}

  \label{fig:qual_comparison}
  \vspace{-0.3cm}
\end{figure}

Table \ref{tab:compare_scalability_reliability} shows the timing results and the success rate of all pair-wise transitions from each strategy when accommodating the motion vocabulary. As expected, strategies that require an additional training process to accommodate the expanded motion vocabulary (Motion Imitation and Gating Network) require a substantial compute time. Even worse, the re-training process may disrupt and alter existing motions. The lower success rate of transitions between Trot and Canter after adding the Jump motion further highlights this problem. In contrast, strategies that do not require any additional training processes preserve the existing motions while achieving better scalability with future vocabulary expansion. However, since the Value Function does not consider the transition outcomes, it generates unstable transitions, which often cause the character to fall. Our approach needs to populate the transition tensor to identify good transitions, which does not update the controller. Therefore, we maintain a competitive compute time compared to the Value Function strategy while offering more reliable transitions highlighted by higher success rates. Figure \ref{fig:qual_comparison} visually demonstrates how each strategy performs.

\npdecimalsign{.}
\nprounddigits{2}
\setlength\tabcolsep{2pt} 
\newcolumntype{?}{!{\vrule width 0.6pt}}
\begin{table}[t!]
    \caption{The performance of each strategy in accommodating iteratively growing motion vocabulary. Trot, Canter, and Jump motions are denoted as T, C, and J respectively. Our approach achieves a competitive training time and produces the highest success rates for all pair-wise transitions. }
    \centering 
    \begin{tabular}{ ?l|l?r?r|r|r|r|r|r? } 
       \Xhline{2\arrayrulewidth}
        \multirow{2}{*}{\textbf{Strategy}} & \multirow{2}{*}{\textbf{Vocab.}} & \multicolumn{1}{c?}{\textbf{Train}}    & \multicolumn{6}{c?}{\textbf{Transition (Src/Dst)}}\\ \cline{3-9}
                                  &                   & (hours)     & \multicolumn{1}{c|}{T/C} & \multicolumn{1}{c|}{T/J} & \multicolumn{1}{c|}{C/T} & \multicolumn{1}{c|}{C/J} & \multicolumn{1}{c|}{J/T} & \multicolumn{1}{c?}{J/C}\\
        \Xhline{2\arrayrulewidth}
       
        Motion  & \text{T+C}         & 155.96 & 0.69 & ~       & 0.94   & ~        & ~        & ~       \\
         Imitation          & \text{T+C+J} & 300.00  & 0.16 & 0      & 0.34   & 0        & 0.56 & 0.14 \\
        
        \Xhline{2\arrayrulewidth}
        
        Gating  & \text{T+C}        &  71.21  & \textbf{1.00} & ~      & 0.83     & ~    & ~       & ~        \\
        Network           & \text{T+C+J}      & 125.26  & 0.90 & 0.82   & 0.57   & 0.88     & 0.51  & 0.74               \\
       
        \Xhline{2\arrayrulewidth}
       
        Value & \text{T+C}         &  \textbf{47.49}  & 0.74 & ~      & 0.38 & ~    & ~       & ~         \\
        Function       & \text{T+C+J}      &  \textbf{62.03} & 0.74 & 0.72    & 0.38 & 0.50 & 0.49    & 0.45      \\
        
        \Xhline{3\arrayrulewidth}

        \textbf{TMT} & \text{T+C}        &  52.07 & \textbf{1.00} & ~       & \textbf{0.95} & ~        & ~        & ~        \\
        \textbf{(Ours)}           & \text{T+C+J}      &  68.91 & \textbf{1.00} & \textbf{0.96} & \textbf{0.95} & \textbf{0.94} & \textbf{0.96} & \textbf{0.91} \\
        
        \Xhline{2\arrayrulewidth}
       
    \end{tabular}
    \label{tab:compare_scalability_reliability}
    \vspace{-0.2cm}
    
\end{table}

\npnoround

\subsection{Applicability to Different Characters}
Our approach identifies good transition strategies through careful examination of the transition timing and the character's pose. It applies to various characters, such as 80 DoF quadruped and 34 DoF biped. Moreover, it can also accommodate controllers with different architectures and training procedures, enabling us to use publicly available humanoid controllers, like one developed by \citep{peng2018deepmimic} as our framework's template controllers. Figure \ref{fig:showcase_result} demonstrates the transition tensor's capability to identify intricate strategies to perform challenging sequences that require precise timing, such as quickly decelerating from cantering to the quadruped's sitting position and smoothly connecting Jog-Run-Roll for the biped. More results are available in the supplementary video.

\vspace{-0.3cm}
\section{Discussion and Conclusion}
We introduce the transition motion tensor (TMT), a data-driven approach that unifies multiple template controllers into a single coherent controller using novel transitions and careful temporal analysis. The design choice on avoiding the additional training process when including more motions is crucial for scalability, as the process becomes more tractable and independent. Furthermore, by considering the property of the motions, it becomes relatively straightforward to apply our approach for controllers with different network architectures, characters, or even training procedures. This capability is highly desirable as it promotes the reusability of previously designed controllers and shifts the focus of future research directions towards adopting more challenging and intricate motions. In the future, we would like to consider other parameters such as the control and blending parameters, as well as various stylistic properties to further enrich the transition motions.

\begin{acks}
We thank the anonymous reviewers for their insightful comments and our colleagues at Inventec Corporation for their practical feedback that helped improve this paper.
\end{acks}

\bibliographystyle{ACM-Reference-Format}
\bibliography{main}

\end{document}